\documentclass[lettersize,journal]{IEEEtran}
\usepackage{amsmath,amsfonts}
\usepackage{algorithmic}
\usepackage{algorithm}
\usepackage{array}
\usepackage[caption=false,font=normalsize,labelfont=sf,textfont=sf]{subfig}
\usepackage{textcomp}
\usepackage{stfloats}
\usepackage{url}
\usepackage{verbatim}
\usepackage{graphicx}
\usepackage{cite}
\usepackage{color}
\usepackage{mhchem}
\begin{document}


\title{Subgraph Networks Based Contrastive Learning}


\author{Jinhuan Wang*,
        Jiafei Shao*,
        Zeyu Wang,
        Shanqing Yu,
        Qi Xuan,~\IEEEmembership{Senior Member,~IEEE},
        Xiaoniu Yang

\IEEEcompsocitemizethanks{
\IEEEcompsocthanksitem This work has been submitted to the IEEE for possible publication. Copyright may be transferred without notice, after which this version may no longer be accessible.
\IEEEcompsocthanksitem J. Wang, J. Shao, Z. Wang, and S. Yu are with the Institute of Cyberspace Security, College of Information Engineering, Zhejiang University of Technology, Hangzhou 310023, China. E-mail: jhwang@zjut.edu.cn; 
shaojiafei1129@gmail.com;
Vencent\_Wang@outlook.com; yushanqing@zjut.edu.cn.
\IEEEcompsocthanksitem Q. Xuan is with the Institute of Cyberspace Security, College of Information Engineering, Zhejiang University of Technology, Hangzhou 310023, China, with the PCL Research Center of Networks and Communications, Peng Cheng Laboratory, Shenzhen 518000, China. E-mail: xuanqi@zjut.edu.cn.
\IEEEcompsocthanksitem X. Yang is with the Institute of Cyberspace Security, Zhejiang University of Technology, Hangzhou 310023, China, and also with the Science and Technology on Communication Information Security Control Laboratory, Jiaxing 314033, China. E-mail: yxn2117@126.com.
\IEEEcompsocthanksitem Corresponding author: Qi Xuan.
\IEEEcompsocthanksitem * Co-first author.
}
}




\maketitle

\begin{abstract}
Graph contrastive learning (GCL), as a self-supervised learning method, can solve the problem of annotated data scarcity.
It mines explicit features in unannotated graphs to generate favorable graph representations for downstream tasks. 
Most existing GCL methods focus on the design of graph augmentation strategies and mutual information estimation operations.
Graph augmentation produces augmented views by graph perturbations.
These views preserve a locally similar structure and exploit explicit features. 
However, these methods have not considered the interaction existing in subgraphs.
To explore the impact of substructure interactions on graph representations, we propose a novel framework called subgraph network-based contrastive learning (SGNCL). 
SGNCL applies a subgraph network generation strategy to produce augmented views.
This strategy converts the original graph into an Edge-to-Node mapping network with both topological and attribute features.
The single-shot augmented view is a first-order subgraph network that mines the interaction between nodes, node-edge, and edges.
In addition, we also investigate the impact of the second-order subgraph augmentation on mining graph structure interactions, and further, propose a contrastive objective that fuses the first-order and second-order subgraph information. 
We compare SGNCL with classical and state-of-the-art graph contrastive learning methods on multiple benchmark datasets of different domains.
Extensive experiments show that SGNCL achieves competitive or better performance (top three) on all datasets in unsupervised learning settings. 
Furthermore, SGNCL achieves the best average gain of 6.9\% in transfer learning compared to the best method.
Finally, experiments also demonstrate that mining substructure interactions have positive implications for graph contrastive learning.

\end{abstract}


\section{Introduction}
Many complex systems in the real world are abstractly represented as graphs, such as biological protein networks~\cite{walter2004visualization}, chemical molecular networks~\cite{wale2008comparison}, and social networks~\cite{adamic2003friends, xuan2019self}.
Graph neural networks (GNNs)~\cite{wu2020comprehensive, zhou2020graph,xu2018powerful} can directly and effectively capture structural features in graphs, which learn the contextual relevance of nodes and update node representation by aggregating domain information.
Given the powerful scalability of GNNs, researchers have proposed a large number of related methods based on the GNN layer to solve practical classification or regression problems, such as prediction of drugs’ pharmacological activity~\cite{wang2021multi} or friend recommendation in social networks~\cite{adamic2003friends}.
The vast majority of those methods are implemented in a supervised learning environment~\cite{welling2016semi, velickovic2017graph,li2020distance,hu2019hierarchical}, and they are highly dependent on high-quality and large-scale annotated data.
However, in practical application scenarios, data with accurate annotation is insufficient, expensive, and resource-consuming~\cite{sun2019infograph,hu2019strategies}.
Therefore, learning graph representations in self-supervised settings becomes increasingly necessary.


Graph contrastive learning (GCL)~\cite{you2020graph, wu2021self,peng2020graph,hassani2020contrastive,zhu2021graph,suresh2021adversarial, you2021graph, li2022let }, as a self-supervised learning method, has attracted much attention recently.
GCL constructs positive augmented views similar to the local structure of the original graph and makes their low-dimensional embeddings close to each other in the latent space~\cite{you2020graph,xu2021self}. 
On the contrary, negative augmented views with different origins will stay away from each other.
The training process does not rely on annotated data, which solves the problem of label scarcity.
According to the existing research on GCL, most methods focus the design of the general paradigm system on two parts:
(1) \textit{Graph augmentation strategy}~\cite{wu2021self, zhu2021graph, suresh2021adversarial}.
Graph augmentation mainly constructs augmented views through various graph perturbation strategies, which have the property of local structure similarity.
(2) \textit{Mutual information estimation}~\cite{hassani2020contrastive, zhu2021graph,sun2019infograph}. 
It designs a contrastive objective function under the principle of mutual information maximization (InfoMax)~\cite{velickovic2019deep}.
It aims to maximize the similarity between positive views and minimize the similarity between negative views.
Considering only graph augmentation strategies, the existing GCL methods can be further divided into four categories: feature-based augmentation ~\cite{you2020graph,jin2020self,hu2019strategies,xia2022simgrace}, topology-based augmentation~\cite{wang2021multi,zeng2021contrastive}, sampling-based augmentation~\cite{qiu2020gcc,zeng2021contrastive}, and adaptive-based augmentation~\cite{zhu2021graph, li2022let}.
For a certain GCL method, it may contain one or more graph augmentation strategies.
However, those strategies only focus on exploring the influence of explicit local features on graph representation, without considering the interaction correlations hidden between local subgraphs.

Mining substructural interactions is a challenge in graph contrastive learning. 
To tackle this challenge, we introduce the SubGraph Network model (SGN) into the GCL field.
This model relies on the line graph theory that emerged in the 20th century~\cite{harary1960some,gutman1996topological}.
Extensive research has confirmed the effectiveness of line-graph for complex network analysis~\cite{xuan2019subgraph,jo2021edge,chen2017supervised,cai2021line,choudhary2021atomistic}.
For example, Chen et al.~\cite{chen2017supervised} proposed to augment the GNN model with line-graph and non-backtracking operators. 
Cai et al.~\cite{cai2021line} proposed a line graph neural network model for the link prediction task.
Based on the construction rules of the line graphs,  
Xuan et al.~\cite{xuan2019subgraph} introduced the concept and construction method of the subgraph network, and applied it to graph classification tasks successfully.
In addition, they demonstrate that SGNs can be used to expand the structural feature space since SGNs of different orders can capture different aspects of the graph structure.

\begin{figure}[!t]
	\centering
 \includegraphics[width=1\linewidth]{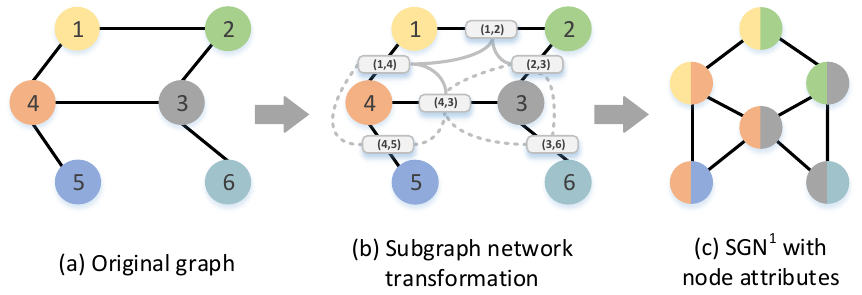}
	\caption{Fundamentals of Subgraph network (SGN) transformation. Specifically, the edges in the original graph are transformed into nodes in SGNs, and the adjacent edges sharing the same node in the original graph will be connected in the SGN. We subsequently extend the rules of SGN and use it as an augmented view.}
	\label{fig:sgn}
\end{figure}

Inspired by this, we revisit the GCL paradigm from the perspective of line-graph theory and subgraph network. As shown in Fig.~\ref{fig:sgn}, the SGN performs transformation operations at the graph topology level. It converts edges into nodes and the pairs of adjacent edges (an opened triangle) into edges.
We develop the idea of SGN to perform line-level subgraph transformations from both graph topology and graph features and integrate them into the GCL, producing a novel contrastive learning framework. 
Specifically, we introduce the \emph{Edge-To-Node} function to generate first-order and second-order SGNs as augmentation views to train the following contrastive model. 
Further, we construct the new contrastive objective function by using the graph representation of these augmented graphs. It requires the original graph and its SGNs to maximize their consistency after the encoder network $\mathcal{T}(\cdot)$.
Formally, $\mathcal{T}(\texttt{G}) \approx \mathcal{T}(\texttt{SGN}(\texttt{G}))$, where $\mathcal{T}(\texttt{G})$ preserves the graph original semantics, and $\mathcal{T}(\texttt{SGN}(\texttt{G}))$ preserves node-node, node-edge, and edge-edge interaction information. 

{\bf Our main contributions are summarized as follows: }
\begin{itemize}
    \item We associate SGN with GCL and propose a new framework, \textbf{S}ub\textbf{G}raph \textbf{N}etwork-based \textbf{C}ontrastive \textbf{L}earning (SGNCL).
    It can capture the interaction information between substructures hidden in the original graph.
    \item In the graph augmentation module, we introduce a subgraph network generation strategy to perform Edge-To-Node transformations on the original graph from graph topology and graph attributes, respectively.
    \item We construct and explore the impact of first-order and second-order SGN on contrastive learning and downstream classification tasks. In addition to single-order SGN-based contrastive loss, we further propose contrastive learning loss based on fused multi-order SGN for simultaneously learning first-order and second-order subgraph feature information.
    \item Extensive experiments demonstrate the effectiveness of our work compared to classical and state-of-the-art contrastive learning methods on multiple benchmark datasets from different domains.
\end{itemize}

The rest of the paper is structured as follows. In Sec.~\ref{sec:related}, we briefly describe the related work on graph augmentation and graph contrastive learning. In Sec.~\ref{sec:method}, we introduced the SGNCL framework. In Sec.~\ref{sec:experiments}, we compare our proposed method with several state-of-the-art baselines in unsupervised learning and transfer learning. Finally, we conclude the paper and highlight some promising directions for future work in Sec.~\ref{sec:conclusion}. 

\section{Related work}\label{sec:related}

\subsection{Subgraphs in Graph Augmentation}

Subgraphs are of great significance to the research of graph mining. In recent years, many subgraph-based methods~\cite{zhang2021nested, you2021identity, yu2020offer} have emerged, which can enhance the capability of graph representation learning by considering subgraph information. 
Some studies are aimed at accelerating the graph augmentation learning strategies by effective subgraph sampling methods~\cite{you2020graph, wang2021sampling, ding2022data}. 
Similarly, Jiao et al.~\cite{jiao2020sub} samples a collection of subgraphs containing the target nodes to enhance the training data for learning the augmented node representation. 
Moreover, Wang et al.~\cite{wang2020graphcrop} provided a subgraph pruning-based graph augmentation method named GraphCrop, which removes the massive noise in graphs and achieves better graph classification performance. 
Most relevant to our approach, Wang et al.~\cite{xuan2019subgraph} utilized graph substructures as the material to generate the subgraph networks of original graphs to serve as the training data for learning an augmented graph classification model. Generally, different from methods based on node and edge modification~\cite{rong2019dropedge, wang2020nodeaug, feng2020graph}, subgraph-based augmentation methods take local structural consistency and global structure dependence into consideration. In this paper, our method, based on the construction of subgraph networks, can preserve the potential structural information to obtain better-augmented graphs for subsequent contrastive learning.

\subsection{Graph Contrastive Learning}
Recent advances in self-supervised learning (SSL)~\cite{grill2020bootstrap, lan2019albert} provide a new breakthrough for solving the problem of label shortage.
As one of the SSL methods, contrastive learning learns explicit features in graphs by comparing the similarities and differences between data and has attracted much attention in recent years.
Research on contrastive learning has focused on the design of graph augmentation and the contrastive objective.
For graph augmentation, augmentation strategies are distinguished into four categories: feature-based, topology-based, sampling-based, and adaptive-based augmentation~\cite{wu2021self}.
For example, GraphCL~\cite{you2020graph} proposes four classic augmentation strategies, including node dropping, edge perturbation, attribute masking, and subgraph.
GCA~\cite{zhu2021graph} considers the prior knowledge from graph topology and semantics for graph sampling and proposes an augmentation strategy based on node centrality measurement.
RGCL~\cite{li2022let} combines GCL with invariant rationale discovery, using a rationale generator to interpret explicit features in graph instances, and adaptively generates augmented views.
These strategies only study the impact of individual substructures on graph representation and do not mine the interaction between substructures.

Regarding the contrastive objective, contrastive learning can compare two views at the same scale or at different scales. Examples of contrastive learning at the same scale are \emph{global-global}~\cite{you2020graph, zeng2021contrastive},\emph{context-context}~\cite{qiu2020gcc} and \emph{local-local}~\cite{yang2022dual}. On the other hand, contrastive learning at different scales includes \emph{global-local}~\cite{hassani2020contrastive}, \emph{global-context}~\cite{sun2019infograph} and \emph{local-context}~\cite{mavromatis2020graph}. In our work, we use subgraph networks with the edge-to-node paradigm of attribution mapping as the augmented views. The graph-level representation of SGNs also contains the context information from the original graph in addition to the global graph information. Therefore, the design of our contrastive objective can be understood as a paradigm of \emph{global-global\&context}.

\begin{figure*}[!t]
	\centering
 \includegraphics[width=1\linewidth]{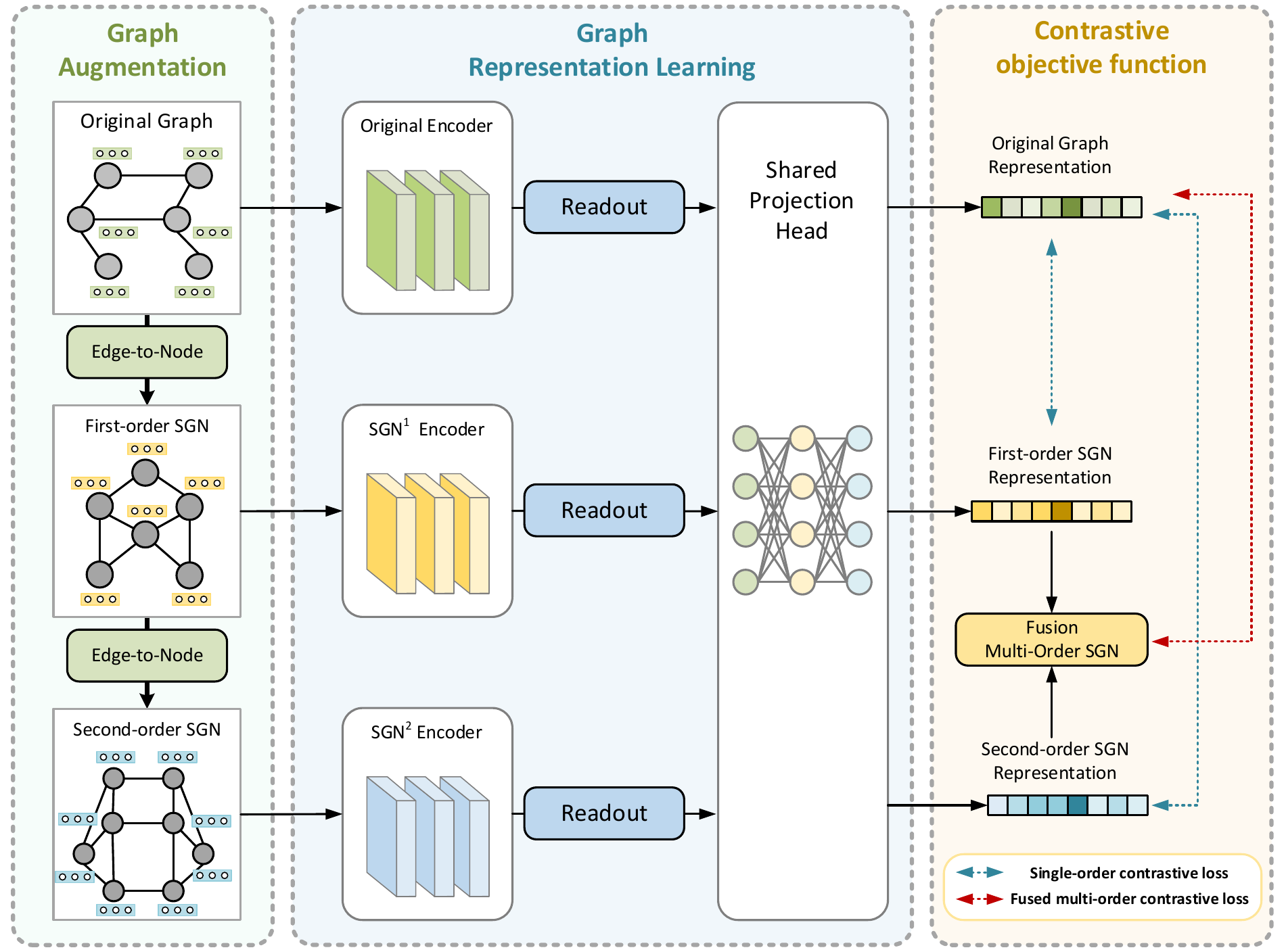}
	\caption{A general framework for SGNCL. In the graph augmentation module, we use an SGN-based strategy to generate subgraph networks (SGNs) in an Edge-to-node fashion with topological and attribute features. Specifically, the first-order SGN mines high-order interaction information in node-node or node-edge; The second-order SGN mines high-order interaction information between subgraphs. These SGNs will serve as augmented views, where original/augmented views with consistent origins are regarded as positive samples, and those views with inconsistent origins are regarded as negative samples. In graph representation learning, all views will be fed into the encoder group to obtain node-level representations. The encoders for different order SGN are independent of each other. Through the readout function, these node-level representations are compressed into graph-level representations. A shared projection head projects representations into a more rigid latent space. In the comparative objective function, we first use the classical function to make the original graph representation close to one of its subgraph network representations in the latent space. Further considering the impact of multi-order SGNs, we propose a fused multi-order contrastive objective function that can approach multi-order subgraph network representations simultaneously.}
	\label{fig:framework}
\end{figure*}

\section{Methodology}\label{sec:method}

Drawing inspiration from line-graph theory and subgraph networks, we introduce SGNCL, a novel framework for contrastive learning that leverages the power of high-order interactions among substructures. Our framework explores the potential of subgraph networks in the field of contrastive learning. In this section, we start with an overview of the entire SGNCL framework, followed by a detailed description of each module.

\subsection{Overview}
The overall framework of SGNCL is shown in Fig. \ref{fig:framework}. The whole framework can be divided into three parts: graph augmentation, graph representation learning, and graph contrastive learning.
\emph{Graph augmentation module}. We use an SGN-based augmentation strategy to generate augmented views in an edge-to-node fashion, such as the first-order SGN ($\texttt{SGN}^{\texttt{1}}$) and the second-order SGN ($\texttt{SGN}^{\texttt{2}}$). These new views can directly display the connection relationship between edges or open triangular motifs.
\emph{Graph representation module}. We provide different encoders for different views. 
These encoders learn graph topology and attribute features to encoder node representations.
Next, these node representations are converted into graph representations by integrating the node information with the readout function.
\emph{Graph contrastive module}. We project graph representations into a low-dimensional space, followed by inter-view comparisons in a stricter latent space. 
GCL requires that the similarity of positive pairs is maximized and the similarity of negative pairs is minimized. 
Therefore, we increase the similarity between the original graph and its different-order SGNs and reduce the similarity between the original graph and SGNs of different origins. 
Further, in order to learn important interaction information from first-order and second-order SGNs simultaneously, we also propose a novel contrastive objective function.
It fuses multi-order information and adjusts the bias through a hyperparameter $\boldsymbol{q}$.
In the next subsections, we will introduce each module in detail.

\subsection{SGN-based Graph Augmentation}
The innovation of graph augmentation in contrastive learning can be summarized into two aspects.
(1) Modification in graph topology, such as node dropping or edge addition/deletion.
These strategies have been further refined under in-depth research, i.e. sampling the original graph based on a certain probability distribution~\cite{hassani2020contrastive,zeng2021contrastive} or processing unimportant substructures based on domain knowledge~\cite{zhu2021graph, li2022let}.
(2) Feature augmentation on substructure attributes, such as attribute masking~\cite{jin2020self,hu2019strategies} or perturbation encoder adding noise~\cite{xia2022simgrace}, etc.
Existing augmentation strategies only explore the influence of local structural similarity on graph representation, but do not explore the high-order interactions between substructures in the graph.
In order to address this problem, we introduce SGN as a graph augmentation strategy.
Under the existing SGN rules, we further carry out attribute constraints making the nodes and edges of SGN have practical application significance.

Fig.\ref{fig:sgn} illustrates the basic rules of SGN. Given the original graph $\mathcal{G}_{ori}=(\mathcal{V},\mathcal{E})$, where $\mathcal{V}=\{v_i|i=1,2,3,...,n\}$ is the node set, $\mathcal{E}=\{(v_{j},v_{k})|j,k=1,2,...,n, j\neq k\}$ is the edge set, and let $\mathcal{N}=\{e_i|i=1,2,3,...,m\}$ be a set of relabeling all edges in $\mathcal{E}$. $\wedge = \{(e_a,v_c,e_b)|e_a,e_b\in \mathcal{N}, v_c \in \mathcal{V}\}$ is the set of open triangular motifs.
Then, the transformation process of SGN can be expressed as $\mathcal{G}_{sgn}=\mathcal{F}(\mathcal{G}_{ori})=(\mathcal{V}_s, \mathcal{E}_s)$, where the node set $\mathcal{V}_s$ = $\mathcal{N}$ and edge set $\mathcal{E}_s \subseteq (\mathcal{V}_s\times \mathcal{V}_s)$.
Specifically, the SGN transformation is designed in the line graph mapping fashion. If two adjacent edges in $\mathcal{G}_{ori}$ share a node, the corresponding nodes in $\mathcal{G}_{sgn}$ are naturally connected.
The topology of the first-order SGN $\mathcal{G}_{sgn}^{1}$ carries the edge interaction information of the original graph $\mathcal{G}_{ori}$.
Repeating the above operations on the basis of $\mathcal{G}_{sgn}^{1}$ will generate a second-order SGN $\mathcal{G}_{sgn}^{2}$, whose topology carries the interaction information between the opened triangle motifs in $\mathcal{G}_{ori}$. Considering an SGN transformation at the $l$-th iteration, it can be formulated as, 
\begin{alignat}{2}
\mathcal{G}_{sgn}^{l} &= \mathsf{LineGraph}(\mathcal{G}_{sgn}^{l-1}),
\label{eq:lg}
\end{alignat}
where $\mathcal{G}_{sgn}^{l-1}$ is the SGN of the $(l-1)$-th order and $\mathsf{LineGraph}(\cdot)$ is the concrete representation of $\mathcal{F}(\cdot)$.

The above content excavates the interaction information hidden in $\mathcal{G}_{ori}$ from a topological perspective. In this paper, we further explore the interaction information from an attribute perspective to enrich the subgraph network. Formally,
at the $l$-th iteration of SGN transformation, 
\begin{alignat}{2}
(\mathcal{G}_{sgn}^{l}, \mathbf{X}_{sgn}^{l}, \mathbf{U}_{sgn}^{l}) &= \mathsf{EdgeToNode}(\mathcal{G}_{sgn}^{l-1}, \mathbf{X}_{sgn}^{l-1}, \mathbf{U}_{sgn}^{l-1}),
\label{eq:e2n}
\end{alignat}
where $\mathcal{G}_{sgn}^{0}$ = $\mathcal{G}_{ori}$ = $(\mathcal{V},\mathcal{E})$, $\mathbf{X}_{sgn}^{0}$ = $\mathbf{X}_{ori} \in \mathbb{R}^{|\mathcal{V}|\times d}$ is the initial node features, and $\mathbf{U}_{sgn}^{0}$ = $\mathbf{U}_{ori} \in \mathbb{R}^{|\mathcal{E}|\times r}$ is the initial edge features. Specifically, the input of $l$-th iteration is $\mathcal{G}_{sgn}^{l-1}$ = $(\mathcal{V}^{l-1},\mathcal{E}^{l-1})$, $\mathbf{X}_{sgn}^{l-1}\in \mathbb{R}^{|\mathcal{V}^{l-1}|\times d^{l-1}}$, and $\mathbf{U}_{sgn}^{l-1} \in \mathbb{R}^{|\mathcal{E}^{l-1}|\times r^{l-1}}$.
And, the label refinement operation is defined by,
\begin{alignat}{2}
\mathcal{N}^{l-1} &= \mathsf{HASH}(\mathcal{E}^{l-1}), 
\label{eq:hash}
\end{alignat}
where $\mathsf{HASH}(\cdot)$ arranges the collection elements in dictionary order and relabels them. 

According to the basic rule as shown in Eq.~\ref{eq:lg}, $\mathcal{V}^{l} = \mathcal{N}^{l-1}$, then the SGN of the $l$-th order can be presented as,
\begin{alignat}{2}
\mathcal{G}_{sgn}^{l} &= (\mathcal{V}^{l},\mathcal{E}^{l}) = (\mathcal{N}^{l-1},\mathcal{E}^{l}),\\
\mathbf{X}_{sgn}^{l} &\in \mathbb{R}^{|\mathcal{N}^{l-1}|\times d^{l}}, \quad \mathbf{U}_{sgn}^{l} \in \mathbb{R}^{|\mathcal{E}^{l}|\times r^{l}}, 
\label{eq:l}
\end{alignat}
where $\mathbf{X}_{sgn}^{l}$ and $\mathbf{U}_{sgn}^{l}$ are the updated node and edge features of $\mathcal{G}_{sgn}^{l}$. 

For different domains of graph datasets, how to strategically generate data augmentations while keeping the intrinsic attribute matters. 
Specifically, we design a general attribute generation strategy for different simulation scenarios.
Given $\mathbf{X}_{sgn}^{l-1}$ = $[\mathbf{x}_1^{l-1}, \cdots, \mathbf{x}_n^{l-1}]^{T}$ and $\mathbf{U}_{sgn}^{l-1}$ = $[\mathbf{u}_1^{l-1}, \cdots, \mathbf{u}_m^{l-1}]^{T}$ where $\mathbf{x}_n^{l-1}$ and $\mathbf{u}_m^{l-1}$ are the feature vectors of $v_n \in \mathcal{V}^{l-1}$ and $e_m \in \mathcal{N}^{l-1}$, respectively,
\begin{alignat}{2}
\mathbf{X}_{sgn}^{l} &= \mathsf{E}_{(v_i,v_j) \in\mathcal{E}^{l-1}}(\mathsf{CONCAT}(\mathbf{x}_i^{l-1},\mathbf{x}_j^{l-1})), \label{eq:concat1}\\
\mathbf{U}_{sgn}^{l} &= 
\mathsf{E}_{(e_a,v_c,e_b) \in \wedge^{l-1}}(\mathsf{ENCODE}(\mathbf{u}_a^{l-1},\mathbf{x}_c^{l-1},\mathbf{u}_b^{l-1})),
\label{eq:concat2}
\end{alignat}
where $\wedge^{l-1}$ is the collection of open triangular motifs of "edge-node-edge" in $\mathcal{G}_{sgn}^{l-1}$, $e_a$, $e_b \in \mathcal{N}^{l-1}$, $v_c\in\mathcal{V}^{l-1}$ is the shared node of $e_a$ and $e_b$.
$\mathsf{E}_{s}(\cdot)$ is a vertical splicing function with the constraint condition $s$.
$\mathsf{CONCAT}{(\cdot)}$ can horizontally splice the node attribution at the previous layer to generate the node attributes at the next layer, $\mathsf{ENCODE}{(\cdot)}$ can encode the node and edge attributes at the previous layer to the edge attributes at the next layer. It can be designed flexibly according to different professional knowledge backgrounds, such as the simplest "concat" operation.
Finally, we can obtain $\mathcal{G}_{sgn}$, $\mathbf{X}_{sgn}$, and $\mathbf{U}_{sgn}$ of different orders through the function $\mathsf{EdgeToNode}(\cdot)$, which will be regarded as the structural and attribution information of the augmentation view.


According to Eq.~\ref{eq:concat1} and Eq.~\ref{eq:concat2}, the node and edge attributes in the augmented view will be updated as the attribute combination and concatenation of node and edge in the input graph. This will meet the requirement of the message-passing network~\cite{gilmer2017neural} for analyzing node attributes, edge attributes, and adjacency relations. Meanwhile, the well-designed attributes constraint rules are conducive to capturing the intrinsic property and mining the formation mechanism of real-world networks. For example, for social networks, we customize the node degree values as node attributes and construct attribute concatenation to study the structural pattern of different social networks via local degree connection rules. For chemical molecules, the actual meaning of node attributes is atomic properties including atom type and atom chirality, and the edge attributes include the bond type and bond direction. 
By constructing attribute concatenation, we can investigate the impact of molecular local connection patterns on molecular property prediction tasks via attribute combination.

\begin{figure*}[!t]
	\centering
        \includegraphics[width=1\linewidth]{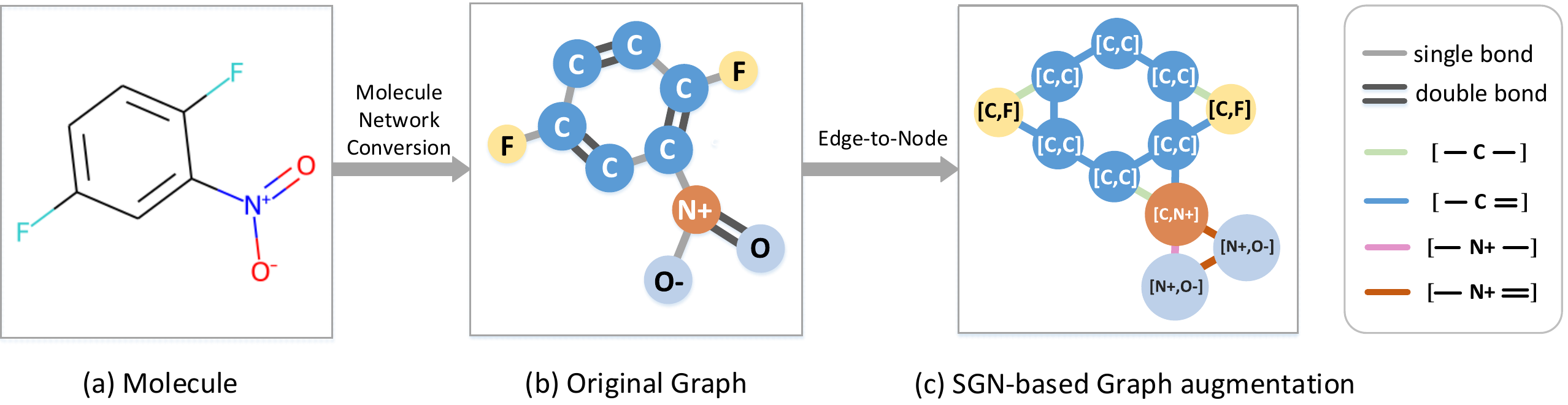}
	\caption{SGN-Based Graph augmentation process in the transfer learning setting. For a molecule, we transform it into a graph, where nodes contain atom-related information and edges contain bond-related information. We generate its first-order SGN topology based on SGN rules. Further, we expand its node attributes and edge attributes. The node attributes are updated to the combination of atom types of linked node pairs in $G_{ori}$, and the edge attributes are updated to the combination of ' bond-central atom-bond' of the triangular motif in $G_{ori}$.
	}
	\label{fig:e2n}
\end{figure*}

Taking a molecular graph as an example, we illustrate the generation progress of our graph augmentation. As shown in Fig.~\ref{fig:e2n} (a), a simple molecule is composed of atoms (nodes) and bonds (edges). 
We convert the molecule into a graph and generate $\texttt{SGN}^{1}$ as the augmentation view according to the Edge-to-Node mapping shown in Eq.~\ref{eq:e2n}. 
The node attributes of the augmented graph are updated to the combination of atom types of node pairs, such as [$\mathsf{C}$,$\mathsf{C}$], [$\mathsf{C}$,$\mathsf{F}$], [$\mathsf{C}$,$\mathsf{N}_{+}$], and [$\mathsf{N}_{+}$,$\mathsf{O}_{-}$]. Further, we update edge attributes to the combination of "\emph{bond-atom-bond}", such as [$\boldsymbol{-}$$\mathsf{C}$$\boldsymbol{-}$], [$\boldsymbol{-}$$\mathsf{C}$$\boldsymbol{=}$], [$\boldsymbol{-}$$\mathsf{N}_{+}$$\boldsymbol{-}$], and [$\boldsymbol{-}$$\mathsf{N}_{+}$$\boldsymbol{=}$]. 
Similarly, the augmented graph $\texttt{SGN}^{2}$ will be obtained after an Edge-to-Node mapping of Fig.~\ref{fig:e2n} (c). It will produce more abundant interaction combinations for node attributes and edge attributes, which are significant for learning a comprehensive graph-level representation. 

\subsection{Graph Representation Learning}

Firstly, we introduce a graph encoder $f_{ori}(\cdot)$ to extract the node representation $\mathbf{h}_{ori}$ for the original graph,
\begin{alignat}{2}
\mathbf{h}_{ori} &= f_{ori}(\mathcal{G}_{ori}, \mathbf{X}_{ori}, \mathbf{U}_{ori}), \label{eq:encode0}
\end{alignat}
According to Eq.~\ref{eq:e2n}, Eq.~\ref{eq:concat1}, and Eq.~\ref{eq:concat2}, the original graph $\mathcal{G}_{ori}$ undergoes the twice graph augmentation operations to generate two related views.
The first conversion operation generates the first-order subgraph network $\mathcal{G}_{sgn}^{1}$, and the second conversion operation generates a second-order subgraph network $\mathcal{G}_{sgn}^{2}$.
In contrast to the original graph, $\mathcal{G}_{sgn}^{1}$ and $\mathcal{G}_{sgn}^{2}$ present topological structures of different sizes and carry abundant attributes of different dimensions.
Therefore, we use independent graph encoders to process different views separately, and their configurations are basically the same except for the input dimension. $\mathbf{h}_{*}$ is a low-dimensional node representation.
%
\begin{alignat}{2}
\mathbf{h}_{sgn1} &= f_{sgn}^{1}(\mathcal{G}_{sgn}^{1}, \mathbf{X}_{sgn}^{1}, \mathbf{U}_{sgn}^{1}), \label{eq:encode1}\\
\mathbf{h}_{sgn2} &= f_{sgn}^{2}(\mathcal{G}_{sgn}^{2}, \mathbf{X}_{sgn}^{2}, \mathbf{U}_{sgn}^{2}), \label{eq:encode2}
\end{alignat}
where $f_{sgn}^{1}$ and $f_{sgn}^{2}$ are encoders for first-order and second-order SGNs, respectively. 
Graph contrastive learning does not apply any constraint to the graph encoder architecture. 
In order to show the best effect, we choose graph isomorphism network (GIN) as our base graph encoder. 
GIN can encode positive pairs into the same embedding and capture the similarity of the graph structure. Next, we use the readout function~\cite{atwood2016diffusion,jo2021edge} to compress node representations into a graph representation $\mathbf{H}$, 
\begin{equation}
\mathbf{H}_*= ||^{K}_{k=1}{(\mathsf{READOUT}(\mathbf{h}_*^{k}))},
\label{eq:readout}
\end{equation}
where $*$ indicates the input view modes, such as $ori$, $sgn_1$ and $sgn_2$. $\mathsf{READOUT}(\cdot)$ is the readout function, which is usually set to classic pooling operations, such as ave-pooling, max-pooling, and sum-pooling.
$||$ is the concatenation operation for all $K$ layer graph representations.

Furthermore, a non-linear transformation $p(\cdot)$ named shared projection maps the graph representation of different views into another latent space where the contrastive loss is applied.
This behavior enforces the mutual information between views to a stricter lower bound~\cite{chen2020simple}.
Its projection process is described as follows,
\begin{alignat}{2}
\mathbf{z}_{ori};\ \mathbf{z}_{\texttt{SGN}}^{\texttt{1}};\ \mathbf{z}_{\texttt{SGN}}^{\texttt{2}} &= p\ (\mathbf{H}_{ori};\ \mathbf{H}_{sgn1};\ \mathbf{H}_{sgn2}), \label{eq:projection}
\end{alignat}
where $p(\cdot)$ consists of a three-layer MLP with $L_2$ normalized outputs. $\mathbf{z}_*$ is the graph representation vector in the latent space.
The model will use the projection head in the pre-training process, but it will discard the projection head when fine-tuning the model for downstream classification tasks.

\subsection{Contrastive Objective Function}
In our SGNCL framework, a contrastive objective function is set to compare the projection vectors from the original graph and augmented views, namely $z$ and $\tilde{z}$. Based on previous research~\cite{you2020graph}, we use the normalized temperature-scaled cross-entropy loss (NT-Xent)~\cite{sohn2016improved, oord2018representation}. 
During SGNCL training, a mini-batch of $I$ original graphs is randomly sampled. 
We perform the SGN transformation in an edge-to-node fashion on mini-batch datasets for contrastive learning.

\textbf{Single-order contrastive loss}. For contrastive learning of single-order SGN, we set two modes including $\mathcal{G}_{sgn}^{1}$ vs $\mathcal{G}_{ori}$ (SGNCL) and $\mathcal{G}_{sgn}^{2}$ vs $\mathcal{G}_{ori}$ (SGNCL-v2).
Both modes satisfy the following contrastive objective function on graph $\mathcal{G}_i$, 
\begin{equation}
    \mathcal{L}_{sim}(\mathcal{G}_i) = -\mathsf{log}(\frac{\exp{(\mathsf{sim}(z_i,\tilde{z_i})/\tau)}}{\sum_{i=1,j\neq i}^I{\exp{(\mathsf{sim}(z_i,\tilde{z_j})/\tau))}}}),
\label{eq:sgn_loss}
\end{equation}
where graph representations with the same origin $\mathcal{G}_i$ are regarded as the positive pair, such as $z_i$ and $\tilde{z_i}$,
graph representations with different origins $\mathcal{G}_i$ and $\mathcal{G}_j$ are regarded as negative pairs, such as $z_i$ and $\tilde{z_j}$,
$\tau$ is the temperature parameter, $\mathsf{sim}(z,\tilde{z})$ represents the cosine similarity function.
It encourages graph representations to be consistent with their augmentation projections that reflect internal interactions while broadening divergence from other graphs.

\textbf{Fused multi-order contrastive loss}. The interaction information fed back by the first-order and second-order SGNs is different, and we expect to further optimize the contrastive loss so that it can learn richer interaction information.
Thus, we propose a fused multi-order contrastive objective function (SGNCL-FU).
Our goal is to make the original graph approach the projection vectors of the first-order and second-order SGNs simultaneously in the latent space.
Further, in order to obtain greater influence weight for salient information in fused multi-order SGNs, we set an adjustable hyperparameter $\boldsymbol{q}$.
It controls the tradeoff between first-order and second-order, 
\begin{equation}
    \mathcal{S}_{sgn*} = \frac{\exp{(\mathsf{sim}(z_i,\tilde{z}^*_i/\tau)}}{\sum_{j=1,j\neq i}^I{\exp{(\mathsf{sim}(z_i,\tilde{z}^*_j)/\tau))}}}, 
\label{eq:fusion_sim}
\end{equation}
\begin{equation}
    \mathcal{L}_{fuse}(\mathcal{G}_i) = -\mathsf{log}( \boldsymbol{q} \times \mathcal{S}_{sgn1} + (1-\boldsymbol{q}) \times \mathcal{S}_{sgn2}), \\
\label{eq:fusion_loss}
\end{equation}
where $*$ indicates the order of SGNs, $\mathcal{S}$ indicates the ratio of positive and negative similarities.

\section{Experiments}\label{sec:experiments}
In this section, we conduct extensive experiments to investigate the performance of our framework on graph classification tasks.
First, we evaluate the performance of SGNCL against state-of-the-art competitors.
Related experiments set up two environment modes: unsupervised learning and transfer learning.
Second, we design ablation experiments to analyze the influence of our contrastive loss with different-order subgraph networks on model performance.
Finally, we perform a hyperparameter sensitivity analysis on the contrastive objective function of fused multi-order SGN.

\subsection{Experimental Setup}
\textbf{Datasets}. To verify the effectiveness of our model, we select eight datasets from the benchmark TUdataset~\cite{morris2020tudataset} for unsupervised learning, including five small molecules \& proteins datasets, MUTAG, PTC-MR, NCI1, PROTEINS, and DD, and three social networks datasets, REDDIT-BINARY, IMDB-BINARY, and IMDB-MULTI. 
Dataset statistics in the unsupervised learning setting are summarized in Table~\ref{tab:1}.

\begin{table}[t]
	\caption{Statistics of datasets for the unsupervised learning task.}
	\label{tab:1}       
	\begin{tabular}{ccccc}
		\hline\hline
		Datasets & Graphs & Classes & Avg.Nodes & Avg.Edges  \\
		\hline
		MUTAG & 188 & 2 & 17.93 &19.79\\
		PTC-MR & 344 & 2 & 14.29 & 14.69\\
		NCI1 & 4,110 & 2 & 29.87 & 32.30\\
		PROTEINS & 1,113 & 2 & 39.06 &72.82\\
		DD & 1,178 & 2 & 284.32 & 715.66\\
		REDDIT-BINARY & 2,000 & 2 & 429.63 & 497.75\\
		IMDB-BINARY & 1,000 & 2 & 19.77 & 96.53\\
        IMDB-MULTI & 1,500 & 3 & 13.00 & 65.93\\
		\hline\hline
	\end{tabular}
\end{table}

Furthermore, we refer to the setting of previous works to evaluate the performance of predicting the chemical properties of molecules in transfer learning for a fair comparison.
We pre-train the backbone GNN model using ZINC-2M, a subset of 2 million unlabeled molecules sampled from the ZINC15 database~\cite{sterling2015zinc}.
We set the downstream task objects are eight binary classification datasets in MoleculeNet~\cite{wu2018moleculenet}, and use scaffold split to segment them to simulate the real-world data distribution.
We use these molecular datasets to fine-tune model parameters, making the model performance more inclined to specific classification task requirements.
The specific statistics of datasets for the transfer learning task are summarized in Table~\ref{tab:2}.

\begin{table}[t]
	\caption{Statistics of datasets for the transfer learning task.}
	\label{tab:2}       
	\begin{tabular}{ccccc}
		\hline\hline
		Datasets & Mode & Graphs & Avg.Nodes & Avg.Edges  \\
		\hline
		ZINC-2M & PRE-TRAIN & 2,000,000 & 26.62 & 57.72\\
		BBBP & FINE-TUNE & 2,039 & 24.06 & 51.90\\
		Tox21 & FINE-TUNE & 7,831 & 18.57 & 38.58\\
		ToxCAST &  FINE-TUNE & 8,576 & 18.78 & 38.52\\
		SIDER & FINE-TUNE & 1,427 & 33.64 & 70.71\\
		CLINTOX & FINE-TUNE & 1,477 & 26.15 & 55.76\\
		MUV & FINE-TUNE & 93,087 & 24.23 & 52.55\\
		HIV & FINE-TUNE & 41,127 & 25.51 & 54.93\\
        BACE & FINE-TUNE & 1,513 & 34.08 & 73.71\\
		\hline\hline
	\end{tabular}
\end{table}

{\bf Learning details.} Based on previous research on self-supervised learning~\cite{you2020graph}, we evaluate the performance of SGN-CL on graph classification tasks in unsupervised and transfer settings.

In unsupervised learning, we train the SGNCL model by utilizing the entire dataset without label information. 
According to \cite{hu2019strategies}, we adopt a 3-layer Graph Isomorphism Network (GIN)~\cite{xu2018powerful} with 32-dimensional hidden units as the encoder. 
The obtained graph representation is fed into a downstream SVM classifier for 10-fold cross-validation. 
We set up five random seeds for randomized experiments and report the average 10-fold cross-validation accuracy. 

In transfer learning, we set the pre-training model as a 5-layer GIN with 300 hidden dimensions. 
The general model is trained by SGNCL and the original encoder parameters are extracted, and these parameters are transferred to the GNN model of the downstream task. Next, we fine-tune model parameters with specific biochemical datasets (e.g., BBBP, SIDER, and ClinTox) to make the model more specialized for specific molecular prediction tasks.
Specifically, the ratio used for training/validation/testing in the dataset is \{8:1:1\}. 
Finally, the test set evaluates the final classification performance of the model.

{\bf Baselines}. Under the unsupervised representation learning setting, we employ six graph-level representation learning methods InfoGraph~\cite{sun2019infograph}, GraphCL~\cite{you2020graph}, JOAO~\cite{you2021graph}, MVGRL~\cite{icml2020_1971}, ADGCL~\cite{suresh2021adversarial}, RGCL~\cite{li2022let} as the baselines. 

Besides, we select several state-of-the-art pre-training paradigms in this area as our baselines under the transfer learning setting. In addition to InfoGraph~\cite{sun2019infograph}, GraphCL~\cite{you2020graph}, JOAO~\cite{you2021graph}, ADGCL~\cite{suresh2021adversarial}, RGCL~\cite{li2022let}, we also adopt the original Edge Prediction EdgePred~\cite{hamilton2017inductive} to predict the connectivity of the node
pairs, Attribute Masking strategy AttrMasking~\cite{hu2019strategies} to learn the regularities of the node and edge attributes distributed over graphs, and Context Prediction strategy ContextPred~\cite{hu2019strategies} to explore graph structures.

\begin{table*}[!t]
    \centering
    \renewcommand{\arraystretch}{1.2}
	\caption{In unsupervised learning setting, comparing classification accuracy(\%) on TUDataset with baselines. The top-3 accuracy(\%) for each dataset is emphasized in bold.}
	\label{tab:3}       
	\begin{tabular}
 {p{43 pt}<{\centering}|p{45 pt}<{\centering}p{45 pt}<{\centering}p{45 pt}<{\centering}p{45 pt}<{\centering}p{45 pt}<{\centering}|p{45 pt}<{\centering}p{45 pt}<{\centering}p{45 pt}<{\centering}}
	    \hline \hline
		Method & NCI1 & PROTEINS & PTC-MR & MUTAG & DD & REDDIT-B & IMDB-B & IMDB-M  \\
		\hline
		Infograph~\cite{sun2019infograph} & 76.20±1.06 &  74.44±0.31 & {\bf61.65±1.43}(2) & {\bf89.01±1.13}(2) & 72.85±1.78 & 82.50±1.42 & {\bf 73.03±0.87}(1) & {\bf 49.69±0.53}(3)\\
		GraphCL~\cite{you2020graph} & 77.87±0.41 & 74.39±0.45 & {\bf60.73±2.10}(3) & 86.80±1.34 & {\bf 78.62±0.40}(2) & 89.53±0.84 & 71.14±0.44 & 48.49±0.63\\
		JOAO~\cite{you2021graph} & {\bf78.07±0.47}(3) &  {\bf74.55±0.41}(3) & 58.03±4.81 & 87.35±1.02 & {\bf 77.32±0.54}(3) & 85.29±1.35 & 70.21±3.08 & 48.67±1.48\\
		MVGRL~\cite{icml2020_1971} & 75.17±0.7 & 71.50±0.30 & 60.45±1.70 & 75.40±7.80 & - & 82.00±1.10 & 63.60±4.20 & {\bf 52.18±0.78}(1)\\
            ADGCL~\cite{suresh2021adversarial} & 73.91±0.77 & 73.28±0.46 & 59.06±0.78 & {\bf88.74±1.85}(3) & 75.79±0.87 & {\bf 90.07±0.85}(2) & 70.21±0.68 & 49.04±0.53\\
            RGCL~\cite{li2022let} & {\bf78.14±1.08}(2) & {\bf 75.03±0.43}(1) & 57.32±3,26 & 87.66±1.01 & {\bf 78.86±0.48}(1) & {\bf 90.34±0.58}(1) & {\bf71.85±0.84}(3) & - \\
		\hline
		SGNCL & {\bf 79.84±0.23}(1) & {\bf74.84±0.98}(2) & {\bf 62.96±0.73}(1) & {\bf 89.13±1.91}(1) & 76.36±3.54 & {\bf 89.66±1.06}(3) & {\bf 72.50±0.40}(2) & {\bf 49.89±1.18}(2)\\
		\hline\hline
	\end{tabular}
\end{table*}

\begin{table*}
    \centering
    \renewcommand{\arraystretch}{1.2}
	\caption{In transfer learning setting, comparing ROC-AUC(\%) sores on different molecular property prediction benchmarks with baselines. The top-4 ROC-AUC(\%) sores are emphasized in bold.}
	\label{tab:4}       
	\begin{tabular}{p{53 pt}<{\centering}|p{35 pt}<{\centering}p{35 pt}<{\centering}
	p{35 pt}<{\centering}p{35 pt}<{\centering}p{35 pt}<{\centering}p{35 pt}<{\centering}p{35 pt}<{\centering}p{35 pt}<{\centering}|p{22 pt}<{\centering}p{22 pt}<{\centering}}
	    \hline \hline
		Method & BBBP & Tox21 & ToxCast & Sider & ClinTox & MUV & HIV & BACE &  Avg. & Gain\\
		\hline
		No pre-train & 65.8±4.5 & 74.0±0.8 & 63.4±0.6 & 57.3±1.6 & 58.0±4.4 & 71.8±2.5     & 75.3±1.9 & 70.1±5.4 & 67.0 & -\\
        EdgePred~\cite{hamilton2017inductive} & 67.3±2.4 & {\bf76.0±0.6}(3) & {\bf64.1±0.6}(2) & 60.4±0.7 & 64.1±3.7 & 74.1±2.1 & 76.3±1.0 & {\bf79.9±0.9}(2) & 70.3 & 2.7\\
        AttrMasking~\cite{hu2019strategies} & 64.3±2.8 & {\bf 76.7±0.4}(1) & {\bf 64.2±0.5}(1) & {\bf61.0±0.7}(4) & 71.8±4.1 & 74.7±1.4 & 77.2±1.1 & {\bf79.3±1.6}(4) & 71.2 & 4.2\\
        ContextPred~\cite{hu2019strategies} & 68.0±2.0 & 75.7±0.7 & {\bf63.9±0.6}(4) & 60.9±0.6 & 65.9±3.8 & {\bf75.8±1.7}(2) & 77.3±1.0 & {\bf79.6±1.2}(3) & 70.9 & 3.9\\
        Infograph~\cite{sun2019infograph} & 68.8±0.8 & 75.3±0.5 & 62.7±0.4 & 58.4±0.8 & 69.9±3.0 & {\bf75.3±2.5}(3) & 76.0±0.7 & 75.9±1.6 & 70.3 & 3.3\\
		GraphCL~\cite{you2020graph} & 69.7±0.7 & 73.9±0.7 & 62.4±0.6 
            & 60.5±0.9 & 76.0±2.7 & 69.8±2.7 & {\bf78.5±1.2}(2) & 75.4±1.4 & 70.8 & 3.8\\
		JOAO~\cite{you2021graph} & {\bf70.2±1.0}(3) & 75.0±0.3 & 62.9±0.5 & 60.0±0.8 & {\bf81.3±2.5}(3) & 71.7±1.4 & 76.7±1.2 & 77.3±0.5 & 71.9 & 4.9\\
		ADGCL~\cite{suresh2021adversarial} &{\bf70.0±1.1}(4) & {\bf76.5±0.8}(2) &  63.1±0.7 & {\bf 63.3±0.8}(1) & {\bf79.8±3.5}(4) &   72.3±1.6 & {\bf78.3±1.0}(3) & 78.5±0.8 & 72.7 & 5.7\\
            RGCL~\cite{li2022let} & {\bf71.4±0.7}(2) & 75.2±0.3 & 63.3±0.2 & {\bf61.4±0.6}(2) &  {\bf83.4±0.9}(2) & {\bf 76.7±1.0}(1) & {\bf77.9±0.8}(4) & 76.0±0.8 & 73.2 & 6.2\\
		\hline
		SGNCL & {\bf 71.6±0.8}(1) & {\bf75.8±0.5}(4) & {\bf63.9±0.4}(3) & {\bf61.2±1.1}(3) & {\bf 84.3±2.7}(1)  & {\bf75.2±3.2}(4) & {\bf 78.9±1.3}(1) & {\bf 80.2±1.3}(1) & {\bf 73.9} & {\bf 6.9}\\
		\hline\hline
	\end{tabular}
\end{table*}

\subsection{Graph Classification with Unsupervised Learning}
To evaluate graph classification capability in the unsupervised learning setting, we compare SGNCL with the above five graph contrastive learning methods. 
These methods have innovations in graph augmentation strategies, and the generated views only extract local structural information.
SGNCL focuses on mining the interaction information within sub-graphs by generating views of the edge-to-node structure.
Therefore, we perform first-order SGN contrastive learning on all datasets.
SGNCL represents a model that trains the original encoder using the first-order SGN to generate augmented views for contrastive learning.
It considers the embedding relationship between the original graph and the first-order SGN in the latent space.
As shown in Table~\ref{tab:3}, the results show that SGNCL has excellent graph classification performance in most datasets which is attributed to mining the interactive information between topological structures.
Compared with other baselines, SGNCL ranks first in NCI1, PTC-MR, and MUTAG. 
In social network datasets such as REDDIT-BINARY, IMDB-BINARY, and IMDB-MULTI, this method ranks among the top three.
This phenomenon shows that SGN mining high-order interactions between substructures are beneficial to obtain the best graph representation.



\subsection{Graph Classification with Transfer Learning}
The graph classification task based on transfer learning can evaluate the transferability of the pre-train model.
Considering the topology complexity and SGN generation efficiency of the ZINC-2M, we pre-train a general SGNCL model with the first-order SGN as augmentation views.
In this phase, the learning rate is set to 0.01, and the training epoch is set to \{20, 40, 60, 80, 100\}.
Then the parameters of the original encoder in SGNCL are migrated and configured in the encoder for specific chemical molecule prediction tasks.
Finally, the encoder parameters are fine-tuned with those local small datasets. 
In fine-tuning phase, we repeat the experiment 10 times with different random seeds and obtained the mean and the standard deviation of ROC-AUC sores for each downstream dataset.

The results are presented in Table. \ref{tab:4}. 
Experiments show that SGNCL achieves the best performance on 4 of 8 datasets compared to the previous best schemes. 
In addition, the average ROC-AUC of SGNCL is better than other methods. 
Compared with the existing research, our SGNCL has surpassed the other strong baselines in many cases such as ADGCL and RGCL, with a gain of 6.9\%.
The above results demonstrate that consistent with contrastive learning methods for constructing locally similar topologies, this method which mines interactive information between graph topologies also contributes to model transferability. 

\subsection{ Ablation Study} 

In this section, we conduct ablation experiments to demonstrate the effectiveness of the contrastive objective function for fusing multi-order SGNs.
We perform second-order and fused multi-order SGN contrastive learning on datasets. 
Different from SGNCL, SGNCL-v2 selects the second-order SGN to generate augmented views for contrastive learning.
SGNCL-FU, as a fusion of multi-order SGN contrastive learning, innovates on the contrastive objective function.
It considers the embedding relationship between the original graph and multi-order SGNs in the latent space.
As shown in Table \ref{tab:ablation}, SGNCL, SGNCL-v2, and SGNCL-FU represent contrastive learning with single-order and fused multi-order SGN. 
SGNCL and SGNCL-v2 explore the interaction of different substructures. 
SGNCL focuses on the interaction between 1-hop neighbor nodes, and SGNCL-v2 focuses on the interaction between triangular motifs.
SGNCL-FU is trained under the fusion contrastive objective function, and its output graph representation will consider the latent information in the first-order and second-order SGNs.
The results show that SGNCL-FU outperforms contrastive learning with single-order SGNs.

\begin{table}[!t]
    \centering
    \renewcommand{\arraystretch}{1.2}
	\caption{Ablation study of our proposed SGNCL model.}
	\label{tab:ablation}       
	\begin{tabular}
 {p{45 pt}<{\centering}|p{45 pt}<{\centering}p{45 pt}<{\centering}p{45 pt}<{\centering}p{45 pt}<{\centering}p{45 pt}<{\centering}}
	    \hline \hline
 		Dataset & SGNCL & SGNCL-v2 & SGNCL-FU  \\
		\hline
		NCI1     & {\bf79.84±0.23} & 79.50±0.69 & 79.69±0.64  \\
            PROTEINS & 74.84±0.98 & 74.88±1.31 & {\bf75.31±0.41}  \\
            PTC-RR   & 62.96±0.73 & 62.51±1.68 & {\bf63.26±1.34}  \\   
            MUTAG    & 89.13±1.91 & 89.05±0.83 & {\bf89.56±1.90}  \\
            IMDB-B   & {\bf72.50±0.40} & 71.76±0.34 & 72.16±0.64  \\
		\hline\hline
	\end{tabular}
\end{table}

\subsection{Visualization Analysis}
\begin{figure*}[!t]
	\centering
        \includegraphics[width=1\linewidth]{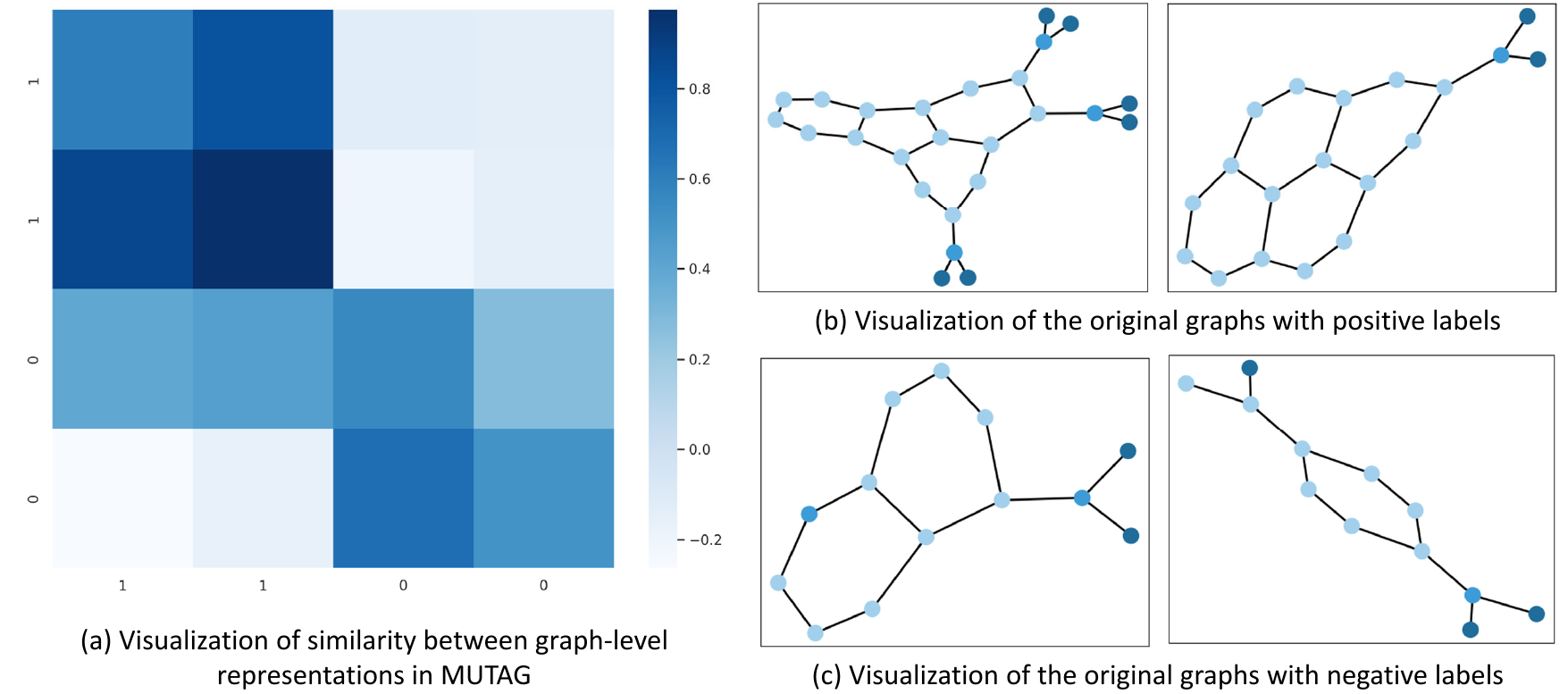}
        \includegraphics[width=1\linewidth]{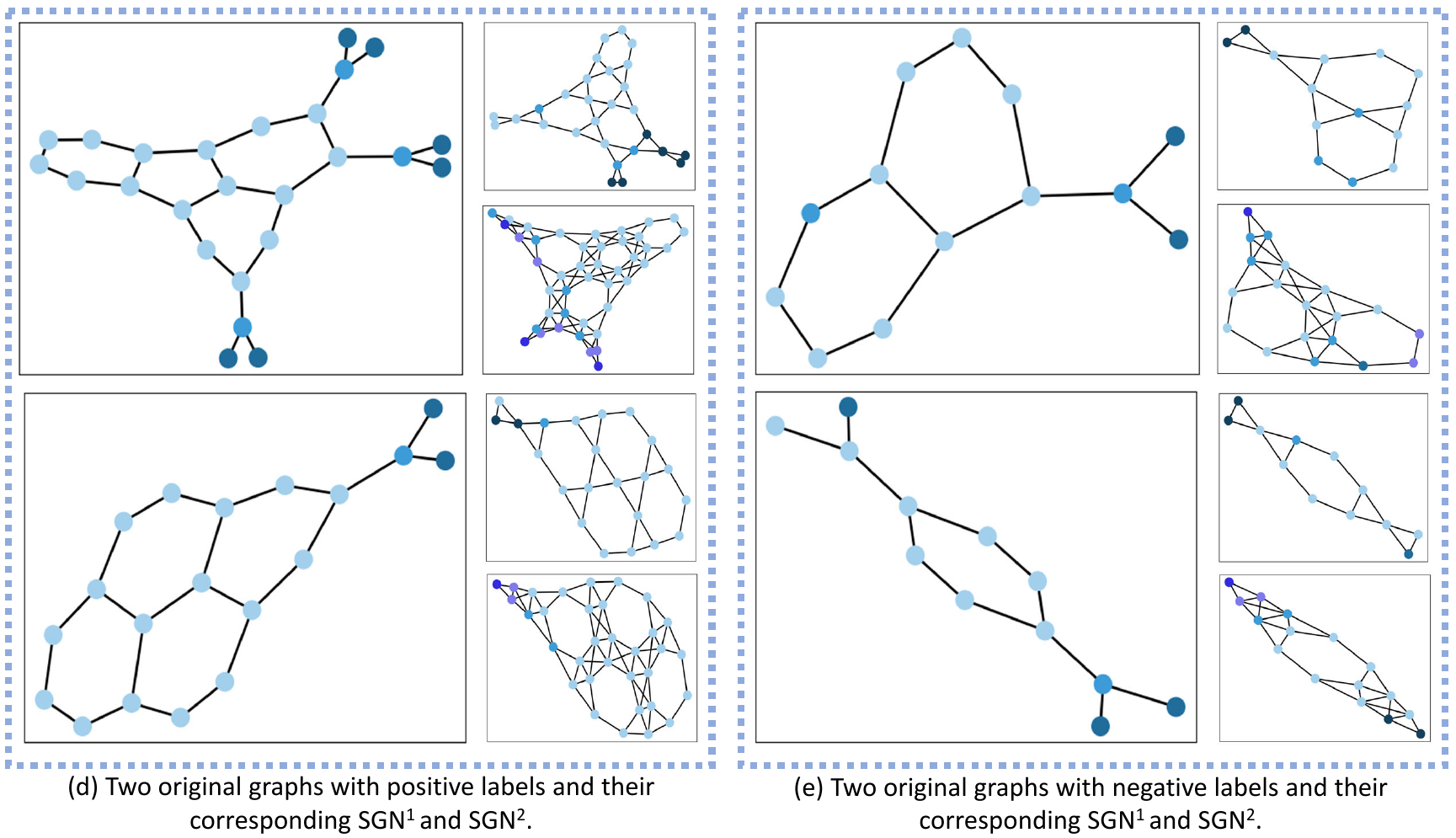}
	\caption{Visualization analysis on MUTAG.
	}
	\label{fig:visual}
\end{figure*}

In order to better explore the augmentation mechanism of SGNCL, we take MUTAG as a case to conduct the visualization analysis.
Fig.~\ref{fig:visual} (a) illustrates the similarity of graph-level representations from any four graphs in MUTAG.
The heat map arranges graph instances horizontally and vertically in the same order. The numbers (1/0) correspond to the actual labels of the instances, while the darker color blocks indicate higher similarity between them. The heat map reveals that instance pairs with the same label tend to be more similar, while those with different labels tend to be less similar. However, there are some exceptions, such as the first two columns of the third row. 
Fig.~\ref{fig:visual} (b) and (c) are visualization results of graph structure for four instances (chemical molecules) with positive and negative labels in (a).
(d) and (e) further dissect the graph instances in (b) and (c), respectively, and visualizes their first-order (upper right) and second-order (lower right) subgraph networks. 

After briefly introducing Fig.~\ref{fig:visual}, we conducted a conjoint analysis of its results. 
Taking (d) as an example, we find that there was a certain commonality between $\texttt{SGN}^{\texttt{1}}$ and $\texttt{SGN}^{\texttt{2}}$ of the graph instances labeled 1. The SGNCL encoder also explores this commonality by mining topology and attribute features in SGNs of different orders. This approach helps improve the classification accuracy in downstream tasks.
When dealing with the unfavorable situation of the first two columns of the third row in (a), we compare the lower graph instance in (d) with the upper graph instance in (d). It is evident that their high similarity is due to the similar topologies of their second-order subgraph networks. 
Hence, in individual cases, only mining the interaction information from the second-order subgraph network may disrupt the classification prediction results. Therefore, this experiment proves that the GCL method, which integrates multi-level SGNs, can obtain more reasonable graph-level representations. This conclusion is also supported by the results of the ablation experiments.

\begin{figure}[!t]
	\centering
\includegraphics[width=1\linewidth]{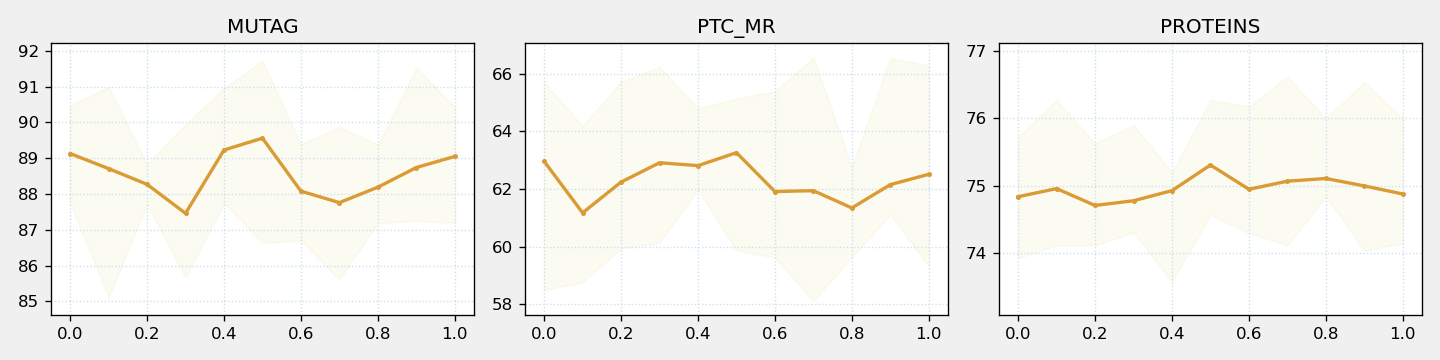}
	\caption{Sensitivity w.r.t. hyperparameter $\boldsymbol{q}$.
	}
	\label{fig:3}
\end{figure}

\begin{figure*}[]
	\centering
 	\includegraphics[width=1\linewidth]{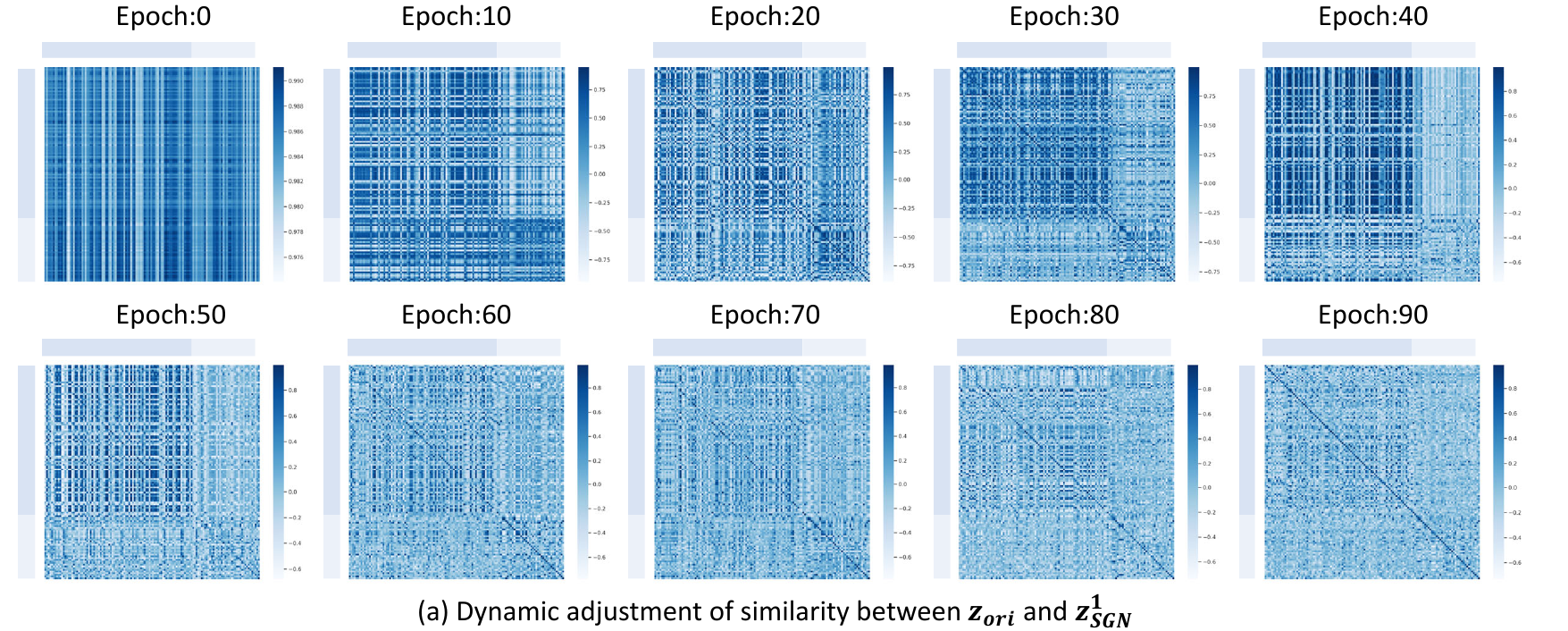}
        \includegraphics[width=1\linewidth]{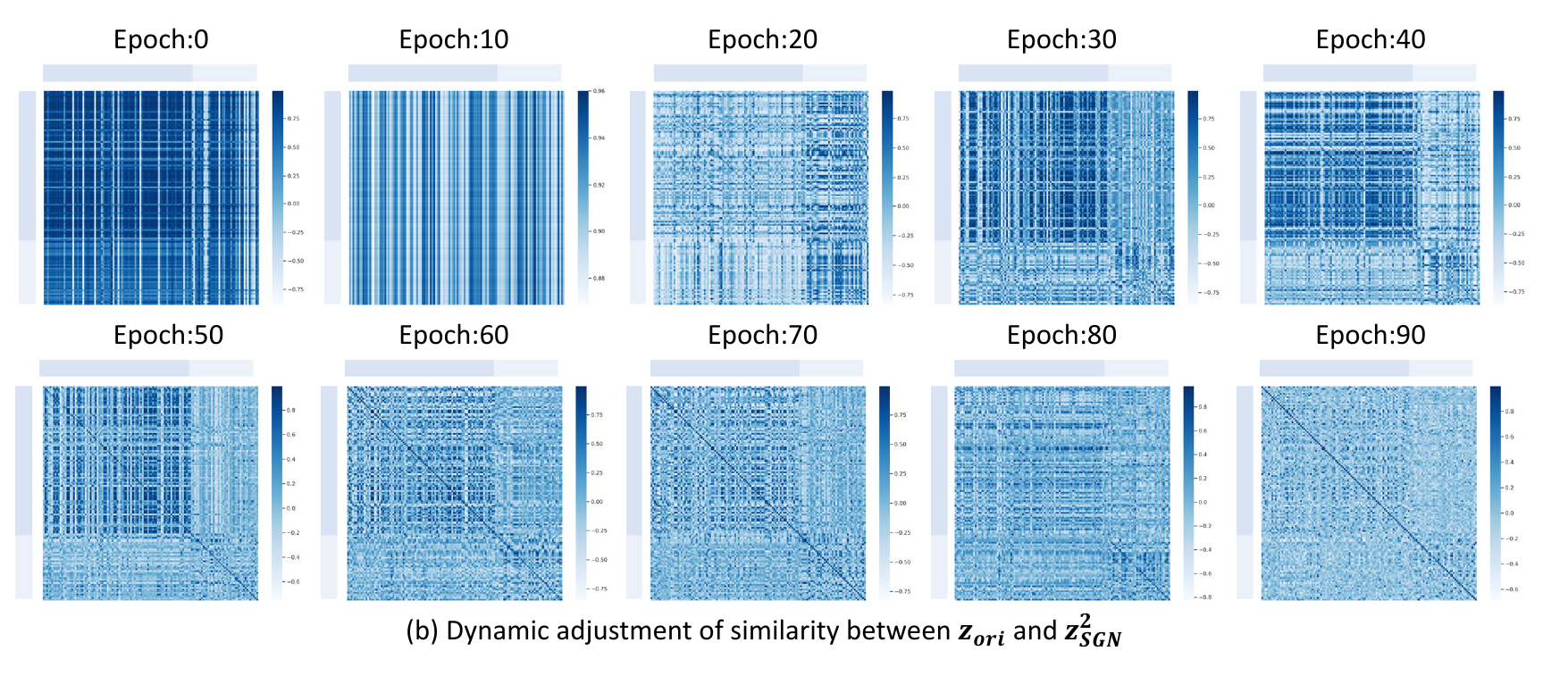}
        \includegraphics[width=1\linewidth]{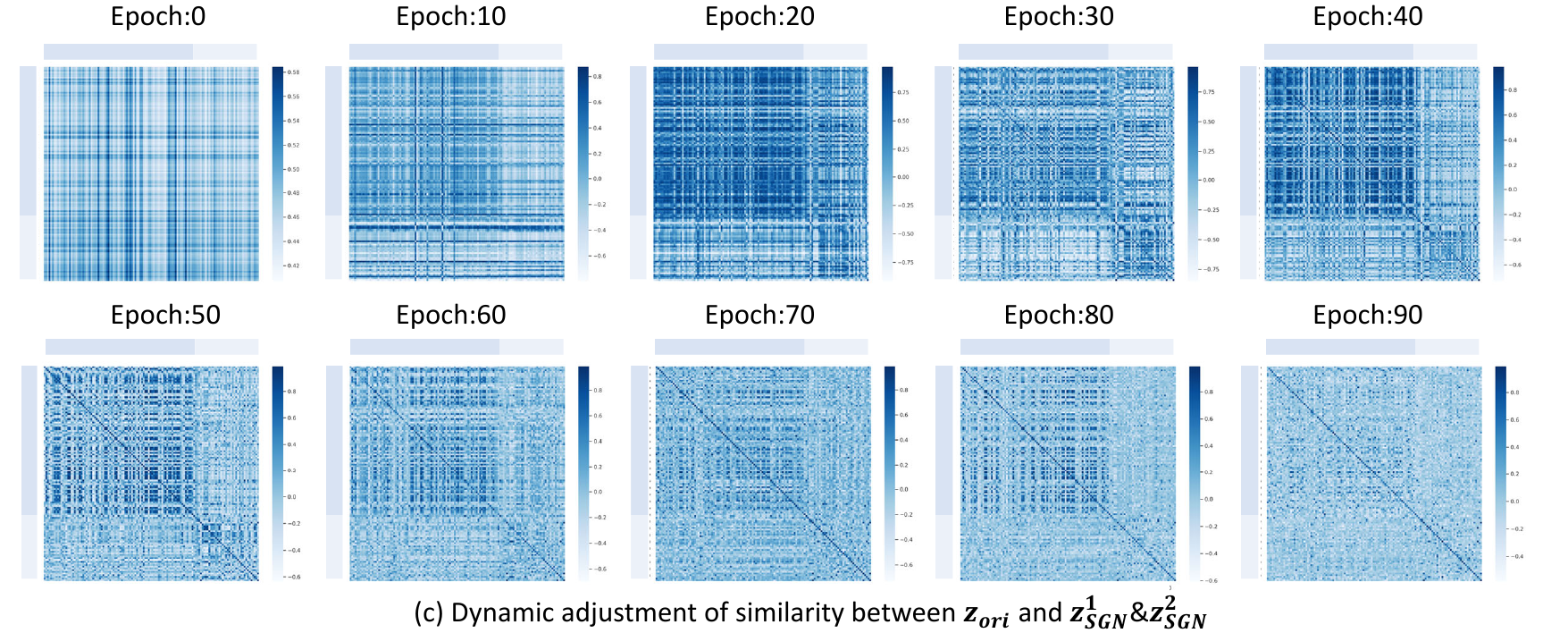}
	\caption{Dynamic adjustment of similarity between views in our method.
	}
	\label{fig:6}
\end{figure*}

\subsection{Hyperparameter Sensitivity}
In this section, we perform sensitivity analysis on the critical hyperparameter $\boldsymbol{q}$ in SGNCL-FU.
Different orders of SGN focus on different aspects of mining interactive information in the original graph topology. Our proposed SGNCL-FU method can effectively capture information from both first-order and second-order SGNs simultaneously. The hyperparameter $\boldsymbol{q}$ in the contrastive objective function plays a critical role in determining the emphasis on mining multi-order SGN information. 

Specifically, as $\boldsymbol{q}$ approaches 0, the resulting graph representation is more similar to the first-order SGN in the latent space, while a higher value of $\boldsymbol{q}$ (approaching 1) emphasizes the second-order SGN. In our experiments on MUTAG, PTC, and PROTEINs datasets, we set the range of $\boldsymbol{q}$ to [0, 1] with a unit interval of 0.1, while keeping other parameters constant. Fig.~\ref{fig:3} illustrates the results, where the average accuracies across all datasets are maximized when $\boldsymbol{q}$ approaches the median value of 0.5. This indicates that the graph representation obtained by the original encoder can significantly benefit the downstream classification task when the embeddings of its first-order and second-order SGNs are equally close to the embedding of the original graph in the latent space.

Additionally, we conduct a hyperparameter sensitivity analysis on the training epochs of the SGNCL model and visualize its results.
We set the value range of epoch to [0, 100] and the experimental interval is 10.
And then, we take MUTAG as an example and randomly select 128 graph instances to visualize the similarity between graph instances and their augmented views by a heat map, as shown in Fig.~\ref{fig:6}.
The horizontal and vertical axes represent the original graph instances and their augmented views, respectively.
In order to make a more intuitive analysis, we put the instance group with label 1 in the front part of the axis, followed by the instance group with label 0.
The boundary between the groups is indicated by the color blocks on the left and top of the heat map.
Note that we used labeled data in this experiment and did not incorporate any data labels during unsupervised learning.
We use dark colors to indicate high similarity between $z_{ori}$ and $z_{sgn}^{*}$, and light colors to indicate low similarity.
Fig.~\ref{fig:6} (a), (b), and (c) correspond to the results obtained using SGNCL, SGNCL-v2, and SGNCL-FU, respectively.

We observe that the similarity between the SGNs of graph instances with the same label increased as the number of training epochs increased. This suggests that there is a certain commonality in the SGNs of graph instances with the same label, which we exploit to help Contrastive Learning extract effective features. Furthermore, the accuracy was highest when the epoch is 40. However, as the epoch number increased beyond 40, only the diagonal color blocks in the heat map became darker, indicating that the SGNCL models tend to overfit and capture individual features instead of common ones. Finally, when comparing Fig.~\ref{fig:6} (c) with Fig.~\ref{fig:6} (a) and Fig.~\ref{fig:6} (b), we found that Fig.~\ref{fig:6} (c) outperformed the other two plots at the epoch of 40, with significantly darker and denser color blocks under the same label.

\section{Conclusion}\label{sec:conclusion}
We revisit the augmentation paradigm in GCL from the perspective of subgraph networks and propose a novel graph contrastive learning framework that mines the latent interactive information in the original graph topology.
This method generates subgraph networks as augmented views in an edge-to-node fashion with node-node, node-edge, and edge-edge interaction attributes.
Further, we apply first-order and second-order SGNs to mine the interaction information between subgraphs.
To simultaneously mine different interaction information, we also propose a novel contrastive objective function for fusing multi-order SGNs.
The experimental results show that considering interaction information as a contrastive item is beneficial to obtain a better graph representation, and it can promote the transferability and generalization of the model.

In future work, we plan to further improve the association framework of SGN and GCL.
We envision extending its framework to a wider range of domains.

\section{ACKNOWLEDGMENTS}

This work was supported in part by the Key R\&D Program of Zhejiang under Grant 2022C01018, by the National Natural Science Foundation of China under Grants 61973273 and U21B2001, by the National Natural Science Foundation of China under Grant 62103374, and by the Key R\&D Projects in Zhejiang Province under Grant 2021C01117.

\bibliographystyle{IEEEtran}
\bibliography{sgn}

\begin{thebibliography}{10}
\providecommand{\url}[1]{#1}
\csname url@samestyle\endcsname
\providecommand{\newblock}{\relax}
\providecommand{\bibinfo}[2]{#2}
\providecommand{\BIBentrySTDinterwordspacing}{\spaceskip=0pt\relax}
\providecommand{\BIBentryALTinterwordstretchfactor}{4}
\providecommand{\BIBentryALTinterwordspacing}{\spaceskip=\fontdimen2\font plus
\BIBentryALTinterwordstretchfactor\fontdimen3\font minus
  \fontdimen4\font\relax}
\providecommand{\BIBforeignlanguage}[2]{{%
\expandafter\ifx\csname l@#1\endcsname\relax
\typeout{** WARNING: IEEEtran.bst: No hyphenation pattern has been}%
\typeout{** loaded for the language `#1'. Using the pattern for}%
\typeout{** the default language instead.}%
\else
\language=\csname l@#1\endcsname
\fi
#2}}
\providecommand{\BIBdecl}{\relax}
\BIBdecl

\bibitem{walter2004visualization}
M.~Walter, C.~Chaban, K.~Sch{\"u}tze, O.~Batistic, K.~Weckermann, C.~N{\"a}ke,
  D.~Blazevic, C.~Grefen, K.~Schumacher, C.~Oecking \emph{et~al.},
  ``Visualization of protein interactions in living plant cells using
  bimolecular fluorescence complementation,'' \emph{The Plant Journal},
  vol.~40, no.~3, pp. 428--438, 2004.

\bibitem{wale2008comparison}
N.~Wale, I.~A. Watson, and G.~Karypis, ``Comparison of descriptor spaces for
  chemical compound retrieval and classification,'' \emph{Knowledge and
  Information Systems}, vol.~14, no.~3, pp. 347--375, 2008.

\bibitem{adamic2003friends}
L.~A. Adamic and E.~Adar, ``Friends and neighbors on the web,'' \emph{Social
  networks}, vol.~25, no.~3, pp. 211--230, 2003.

\bibitem{xuan2019self}
Q.~Xuan, X.~Shu, Z.~Ruan, J.~Wang, C.~Fu, and G.~Chen, ``A self-learning
  information diffusion model for smart social networks,'' \emph{IEEE
  Transactions on Network Science and Engineering}, vol.~7, no.~3, pp.
  1466--1480, 2019.

\bibitem{wu2020comprehensive}
Z.~Wu, S.~Pan, F.~Chen, G.~Long, C.~Zhang, and S.~Y. Philip, ``A comprehensive
  survey on graph neural networks,'' \emph{IEEE transactions on neural networks
  and learning systems}, vol.~32, no.~1, pp. 4--24, 2020.

\bibitem{zhou2020graph}
J.~Zhou, G.~Cui, S.~Hu, Z.~Zhang, C.~Yang, Z.~Liu, L.~Wang, C.~Li, and M.~Sun,
  ``Graph neural networks: A review of methods and applications,'' \emph{AI
  Open}, vol.~1, pp. 57--81, 2020.

\bibitem{xu2018powerful}
K.~Xu, W.~Hu, J.~Leskovec, and S.~Jegelka, ``How powerful are graph neural
  networks?'' \emph{arXiv preprint arXiv:1810.00826}, 2018.

\bibitem{wang2021multi}
Y.~Wang, Y.~Min, X.~Chen, and J.~Wu, ``Multi-view graph contrastive
  representation learning for drug-drug interaction prediction,'' in
  \emph{Proceedings of the Web Conference 2021}, 2021, pp. 2921--2933.

\bibitem{welling2016semi}
M.~Welling and T.~N. Kipf, ``Semi-supervised classification with graph
  convolutional networks,'' in \emph{J. International Conference on Learning
  Representations (ICLR 2017)}, 2016.

\bibitem{velickovic2017graph}
P.~Velickovic, G.~Cucurull, A.~Casanova, A.~Romero, P.~Lio, and Y.~Bengio,
  ``Graph attention networks,'' \emph{stat}, vol. 1050, p.~20, 2017.

\bibitem{li2020distance}
P.~Li, Y.~Wang, H.~Wang, and J.~Leskovec, ``Distance encoding: Design provably
  more powerful neural networks for graph representation learning,''
  \emph{Advances in Neural Information Processing Systems}, vol.~33, pp.
  4465--4478, 2020.

\bibitem{hu2019hierarchical}
F.~Hu, Y.~Zhu, S.~Wu, L.~Wang, and T.~Tan, ``Hierarchical graph convolutional
  networks for semi-supervised node classification,'' \emph{arXiv preprint
  arXiv:1902.06667}, 2019.

\bibitem{sun2019infograph}
F.-Y. Sun, J.~Hoffmann, V.~Verma, and J.~Tang, ``Infograph: Unsupervised and
  semi-supervised graph-level representation learning via mutual information
  maximization,'' \emph{arXiv preprint arXiv:1908.01000}, 2019.

\bibitem{hu2019strategies}
W.~Hu, B.~Liu, J.~Gomes, M.~Zitnik, P.~Liang, V.~Pande, and J.~Leskovec,
  ``Strategies for pre-training graph neural networks,'' \emph{arXiv preprint
  arXiv:1905.12265}, 2019.

\bibitem{you2020graph}
Y.~You, T.~Chen, Y.~Sui, T.~Chen, Z.~Wang, and Y.~Shen, ``Graph contrastive
  learning with augmentations,'' \emph{Advances in neural information
  processing systems}, vol.~33, pp. 5812--5823, 2020.

\bibitem{wu2021self}
L.~Wu, H.~Lin, C.~Tan, Z.~Gao, and S.~Z. Li, ``Self-supervised learning on
  graphs: Contrastive, generative, or predictive,'' \emph{IEEE Transactions on
  Knowledge and Data Engineering}, 2021.

\bibitem{peng2020graph}
Z.~Peng, W.~Huang, M.~Luo, Q.~Zheng, Y.~Rong, T.~Xu, and J.~Huang, ``Graph
  representation learning via graphical mutual information maximization,'' in
  \emph{Proceedings of The Web Conference 2020}, 2020, pp. 259--270.

\bibitem{hassani2020contrastive}
K.~Hassani and A.~H. Khasahmadi, ``Contrastive multi-view representation
  learning on graphs,'' in \emph{International Conference on Machine
  Learning}.\hskip 1em plus 0.5em minus 0.4em\relax PMLR, 2020, pp. 4116--4126.

\bibitem{zhu2021graph}
Y.~Zhu, Y.~Xu, F.~Yu, Q.~Liu, S.~Wu, and L.~Wang, ``Graph contrastive learning
  with adaptive augmentation,'' in \emph{Proceedings of the Web Conference
  2021}, 2021, pp. 2069--2080.

\bibitem{suresh2021adversarial}
S.~Suresh, P.~Li, C.~Hao, and J.~Neville, ``Adversarial graph augmentation to
  improve graph contrastive learning,'' \emph{Advances in Neural Information
  Processing Systems}, vol.~34, pp. 15\,920--15\,933, 2021.

\bibitem{you2021graph}
Y.~You, T.~Chen, Y.~Shen, and Z.~Wang, ``Graph contrastive learning
  automated,'' in \emph{International Conference on Machine Learning}.\hskip
  1em plus 0.5em minus 0.4em\relax PMLR, 2021, pp. 12\,121--12\,132.

\bibitem{li2022let}
S.~Li, X.~Wang, A.~Zhang, Y.~Wu, X.~He, and T.-S. Chua, ``Let invariant
  rationale discovery inspire graph contrastive learning,'' in
  \emph{International Conference on Machine Learning}.\hskip 1em plus 0.5em
  minus 0.4em\relax PMLR, 2022, pp. 13\,052--13\,065.

\bibitem{xu2021self}
M.~Xu, H.~Wang, B.~Ni, H.~Guo, and J.~Tang, ``Self-supervised graph-level
  representation learning with local and global structure,'' in
  \emph{International Conference on Machine Learning}.\hskip 1em plus 0.5em
  minus 0.4em\relax PMLR, 2021, pp. 11\,548--11\,558.

\bibitem{velickovic2019deep}
P.~Velickovic, W.~Fedus, W.~L. Hamilton, P.~Li{\`o}, Y.~Bengio, and R.~D.
  Hjelm, ``Deep graph infomax.'' \emph{ICLR (Poster)}, vol.~2, no.~3, p.~4,
  2019.

\bibitem{jin2020self}
W.~Jin, T.~Derr, H.~Liu, Y.~Wang, S.~Wang, Z.~Liu, and J.~Tang,
  ``Self-supervised learning on graphs: Deep insights and new direction,''
  \emph{arXiv preprint arXiv:2006.10141}, 2020.

\bibitem{xia2022simgrace}
J.~Xia, L.~Wu, J.~Chen, B.~Hu, and S.~Z. Li, ``Simgrace: A simple framework for
  graph contrastive learning without data augmentation,'' in \emph{Proceedings
  of the ACM Web Conference 2022}, 2022, pp. 1070--1079.

\bibitem{zeng2021contrastive}
J.~Zeng and P.~Xie, ``Contrastive self-supervised learning for graph
  classification,'' in \emph{Proceedings of the AAAI Conference on Artificial
  Intelligence}, vol.~35, no.~12, 2021, pp. 10\,824--10\,832.

\bibitem{qiu2020gcc}
J.~Qiu, Q.~Chen, Y.~Dong, J.~Zhang, H.~Yang, M.~Ding, K.~Wang, and J.~Tang,
  ``Gcc: Graph contrastive coding for graph neural network pre-training,'' in
  \emph{Proceedings of the 26th ACM SIGKDD International Conference on
  Knowledge Discovery \& Data Mining}, 2020, pp. 1150--1160.

\bibitem{harary1960some}
F.~Harary and R.~Z. Norman, ``Some properties of line digraphs,''
  \emph{Rendiconti del circolo matematico di palermo}, vol.~9, no.~2, pp.
  161--168, 1960.

\bibitem{gutman1996topological}
I.~Gutman and E.~Estrada, ``Topological indices based on the line graph of the
  molecular graph,'' \emph{Journal of chemical information and computer
  sciences}, vol.~36, no.~3, pp. 541--543, 1996.

\bibitem{xuan2019subgraph}
Q.~Xuan, J.~Wang, M.~Zhao, J.~Yuan, C.~Fu, Z.~Ruan, and G.~Chen, ``Subgraph
  networks with application to structural feature space expansion,'' \emph{IEEE
  Transactions on Knowledge and Data Engineering}, vol.~33, no.~6, pp.
  2776--2789, 2019.

\bibitem{jo2021edge}
J.~Jo, J.~Baek, S.~Lee, D.~Kim, M.~Kang, and S.~J. Hwang, ``Edge representation
  learning with hypergraphs,'' \emph{Advances in Neural Information Processing
  Systems}, vol.~34, pp. 7534--7546, 2021.

\bibitem{chen2017supervised}
Z.~Chen, X.~Li, and J.~Bruna, ``Supervised community detection with line graph
  neural networks,'' \emph{arXiv preprint arXiv:1705.08415}, 2017.

\bibitem{cai2021line}
L.~Cai, J.~Li, J.~Wang, and S.~Ji, ``Line graph neural networks for link
  prediction,'' \emph{IEEE Transactions on Pattern Analysis and Machine
  Intelligence}, 2021.

\bibitem{choudhary2021atomistic}
K.~Choudhary and B.~DeCost, ``Atomistic line graph neural network for improved
  materials property predictions,'' \emph{npj Computational Materials}, vol.~7,
  no.~1, pp. 1--8, 2021.

\bibitem{zhang2021nested}
M.~Zhang and P.~Li, ``Nested graph neural networks,'' \emph{Advances in Neural
  Information Processing Systems}, vol.~34, pp. 15\,734--15\,747, 2021.

\bibitem{you2021identity}
J.~You, J.~M. Gomes-Selman, R.~Ying, and J.~Leskovec, ``Identity-aware graph
  neural networks,'' in \emph{Proceedings of the AAAI conference on artificial
  intelligence}, vol.~35, no.~12, 2021, pp. 10\,737--10\,745.

\bibitem{yu2020offer}
S.~Yu, F.~Xia, J.~Xu, Z.~Chen, and I.~Lee, ``Offer: A motif dimensional
  framework for network representation learning,'' in \emph{Proceedings of the
  29th ACM International Conference on Information \& Knowledge Management},
  2020, pp. 3349--3352.

\bibitem{wang2021sampling}
J.~Wang, P.~Chen, B.~Ma, J.~Zhou, Z.~Ruan, G.~Chen, and Q.~Xuan, ``Sampling
  subgraph network with application to graph classification,'' \emph{IEEE
  Transactions on Network Science and Engineering}, vol.~8, no.~4, pp.
  3478--3490, 2021.

\bibitem{ding2022data}
K.~Ding, Z.~Xu, H.~Tong, and H.~Liu, ``Data augmentation for deep graph
  learning: A survey,'' \emph{ACM SIGKDD Explorations Newsletter}, vol.~24,
  no.~2, pp. 61--77, 2022.

\bibitem{jiao2020sub}
Y.~Jiao, Y.~Xiong, J.~Zhang, Y.~Zhang, T.~Zhang, and Y.~Zhu, ``Sub-graph
  contrast for scalable self-supervised graph representation learning,'' in
  \emph{2020 IEEE international conference on data mining (ICDM)}.\hskip 1em
  plus 0.5em minus 0.4em\relax IEEE, 2020, pp. 222--231.

\bibitem{wang2020graphcrop}
Y.~Wang, W.~Wang, Y.~Liang, Y.~Cai, and B.~Hooi, ``Graphcrop: Subgraph cropping
  for graph classification,'' \emph{arXiv preprint arXiv:2009.10564}, 2020.

\bibitem{rong2019dropedge}
Y.~Rong, W.~Huang, T.~Xu, and J.~Huang, ``Dropedge: Towards deep graph
  convolutional networks on node classification,'' \emph{arXiv preprint
  arXiv:1907.10903}, 2019.

\bibitem{wang2020nodeaug}
Y.~Wang, W.~Wang, Y.~Liang, Y.~Cai, J.~Liu, and B.~Hooi, ``Nodeaug:
  Semi-supervised node classification with data augmentation,'' in
  \emph{Proceedings of the 26th ACM SIGKDD International Conference on
  Knowledge Discovery \& Data Mining}, 2020, pp. 207--217.

\bibitem{feng2020graph}
W.~Feng, J.~Zhang, Y.~Dong, Y.~Han, H.~Luan, Q.~Xu, Q.~Yang, E.~Kharlamov, and
  J.~Tang, ``Graph random neural networks for semi-supervised learning on
  graphs,'' \emph{Advances in neural information processing systems}, vol.~33,
  pp. 22\,092--22\,103, 2020.

\bibitem{grill2020bootstrap}
J.-B. Grill, F.~Strub, F.~Altch{\'e}, C.~Tallec, P.~Richemond, E.~Buchatskaya,
  C.~Doersch, B.~Avila~Pires, Z.~Guo, M.~Gheshlaghi~Azar \emph{et~al.},
  ``Bootstrap your own latent-a new approach to self-supervised learning,''
  \emph{Advances in neural information processing systems}, vol.~33, pp.
  21\,271--21\,284, 2020.

\bibitem{lan2019albert}
Z.~Lan, M.~Chen, S.~Goodman, K.~Gimpel, P.~Sharma, and R.~Soricut, ``Albert: A
  lite bert for self-supervised learning of language representations,''
  \emph{arXiv preprint arXiv:1909.11942}, 2019.

\bibitem{yang2022dual}
H.~Yang, H.~Chen, S.~Pan, L.~Li, P.~S. Yu, and G.~Xu, ``Dual space graph
  contrastive learning,'' in \emph{Proceedings of the ACM Web Conference 2022},
  2022, pp. 1238--1247.

\bibitem{mavromatis2020graph}
C.~Mavromatis and G.~Karypis, ``Graph infoclust: Leveraging cluster-level node
  information for unsupervised graph representation learning,'' \emph{arXiv
  preprint arXiv:2009.06946}, 2020.

\bibitem{gilmer2017neural}
J.~Gilmer, S.~S. Schoenholz, P.~F. Riley, O.~Vinyals, and G.~E. Dahl, ``Neural
  message passing for quantum chemistry,'' in \emph{International conference on
  machine learning}.\hskip 1em plus 0.5em minus 0.4em\relax PMLR, 2017, pp.
  1263--1272.

\bibitem{atwood2016diffusion}
J.~Atwood and D.~Towsley, ``Diffusion-convolutional neural networks,''
  \emph{Advances in neural information processing systems}, vol.~29, 2016.

\bibitem{chen2020simple}
T.~Chen, S.~Kornblith, M.~Norouzi, and G.~Hinton, ``A simple framework for
  contrastive learning of visual representations,'' in \emph{International
  conference on machine learning}.\hskip 1em plus 0.5em minus 0.4em\relax PMLR,
  2020, pp. 1597--1607.

\bibitem{sohn2016improved}
K.~Sohn, ``Improved deep metric learning with multi-class n-pair loss
  objective,'' \emph{Advances in neural information processing systems},
  vol.~29, 2016.

\bibitem{oord2018representation}
A.~v.~d. Oord, Y.~Li, and O.~Vinyals, ``Representation learning with
  contrastive predictive coding,'' \emph{arXiv preprint arXiv:1807.03748},
  2018.

\bibitem{morris2020tudataset}
C.~Morris, N.~M. Kriege, F.~Bause, K.~Kersting, P.~Mutzel, and M.~Neumann,
  ``Tudataset: A collection of benchmark datasets for learning with graphs,''
  \emph{arXiv preprint arXiv:2007.08663}, 2020.

\bibitem{sterling2015zinc}
T.~Sterling and J.~J. Irwin, ``Zinc 15--ligand discovery for everyone,''
  \emph{Journal of chemical information and modeling}, vol.~55, no.~11, pp.
  2324--2337, 2015.

\bibitem{wu2018moleculenet}
Z.~Wu, B.~Ramsundar, E.~N. Feinberg, J.~Gomes, C.~Geniesse, A.~S. Pappu,
  K.~Leswing, and V.~Pande, ``Moleculenet: a benchmark for molecular machine
  learning,'' \emph{Chemical science}, vol.~9, no.~2, pp. 513--530, 2018.

\bibitem{icml2020_1971}
K.~Hassani and A.~H. Khasahmadi, ``Contrastive multi-view representation
  learning on graphs,'' in \emph{Proceedings of International Conference on
  Machine Learning}, 2020, pp. 3451--3461.

\bibitem{hamilton2017inductive}
W.~Hamilton, Z.~Ying, and J.~Leskovec, ``Inductive representation learning on
  large graphs,'' \emph{Advances in neural information processing systems},
  vol.~30, 2017.

\end{thebibliography}

\end{document}